\documentclass{article}

\usepackage{axessibility}  

\usepackage{xcolor}
\usepackage{enumitem}

\definecolor{ForestGreen}{RGB}{34,139,34}

\usepackage{amsmath,amssymb}

\usepackage{pythonhighlight}
\usepackage[square,numbers,sort&compress]{natbib}
\usepackage{algpseudocode}
\usepackage[font=footnotesize,labelfont=bf]{caption}
\usepackage[rightcaption]{sidecap}    
\usepackage{array}
\usepackage{multirow}
\usepackage{booktabs}
\usepackage[linesnumbered,ruled,vlined]{algorithm2e}
\usepackage[font=scriptsize,labelfont=bf]{subcaption}
\usepackage[normalem]{ulem}
\usepackage{xparse}
\usepackage{pifont}
\usepackage{bm}
\usepackage{epigraph}

\usepackage{tabularx}
\usepackage{ifthen}
\usepackage{mwe} 
\usepackage{makecell}
\usepackage{color, colortbl}

\usepackage{wrapfig}

\usepackage[normalem]{ulem}

\usepackage[scaled=0.85]{DejaVuSansMono}

\usepackage[pagebackref=true,breaklinks=true,colorlinks,citecolor=ForestGreen,bookmarks=false]{hyperref}

\usepackage{cleveref}
\crefformat{section}{\S#2#1#3}
\crefformat{subsection}{\S#2#1#3}
\crefformat{subsubsection}{\S#2#1#3}
\crefformat{figure}{Figure~#2#1#3}
\crefformat{equation}{Equation~#2#1#3}
\crefformat{table}{Table~#2#1#3}
\crefformat{algorithm}{Algorithm~#2#1#3}

\SetCommentSty{mycommfont}
\SetKwInput{KwInput}{Input}                

\usepackage{microtype}

\setlist{nosep} 

\newlength\savewidth
\newlength\thinwidth

\definecolor{Gray}{gray}{0.93}
\newcolumntype{a}{>{\columncolor{Gray}}c}
\definecolor{RowOffWhite}{gray}{0.93}
\definecolor{RowWhite}{gray}{0.98}
\definecolor{HighlightBlue}{RGB}{230, 235, 247}

\newcommand{\HC}[1]{\ifthenelse{\isodd{#1}}{\rowcolor{highlightRowColor}}{\rowcolor{white}}}

\newcommand{\rg}{\rowcolor{RowOffWhite}}

\usepackage{xspace}
\makeatletter
\DeclareRobustCommand\onedot{\futurelet\@let@token\@onedot}
\def\@onedot{\ifx\@let@token.\else.\null\fi\xspace}

\def\ie{\emph{i.e}\onedot} 
 
\def\etc{\emph{etc}\onedot}

\makeatother

\newcommand{\Scal}{\mathcal{S}}
\newcommand{\Dcal}{\mathcal{D}}
\newcommand{\Lcal}{\mathcal{L}}

\newcommand{\Bcal}{\mathcal{B}}

\newcommand{\x}{\mathbf{x}}
\newcommand{\z}{\mathbf{z}}

\newcommand{\sg}{\mathtt{sg}}

\usepackage{listings}
\usepackage{xcolor}

\definecolor{codegreen}{rgb}{0,0.6,0}
\definecolor{codegray}{rgb}{0.5,0.5,0.5}
\definecolor{codepurple}{rgb}{0.58,0,0.82}
\definecolor{backcolour}{rgb}{1.,1.,1.}

\lstdefinestyle{mystyle}{
    backgroundcolor=\color{backcolour},   
    commentstyle=\color{codegreen},
    keywordstyle=\color{magenta},
    numberstyle=\tiny\color{codegray},
    stringstyle=\color{codepurple},
    basicstyle=\ttfamily\footnotesize,
    breakatwhitespace=false,         
    breaklines=true,                 
    captionpos=b,                    
    keepspaces=true,                 
    numbers=left,                    
    numbersep=5pt,                  
    showspaces=false,                
    showstringspaces=false,
    showtabs=false,                  
    tabsize=2
}

\usepackage[title]{appendix}

\makeatletter
\newcommand{\printfnsymbol}[1]{%
  \textsuperscript{\@fnsymbol{#1}}%
}
\makeatother

\usepackage{palatino}
\begin{document}
\fontfamily{ppl}\selectfont

\title{The Challenges of\\Continuous Self-Supervised Learning} 

\author{Senthil Purushwalkam\thanks{Authors contributed equally.} \quad Pedro Morgado\printfnsymbol{1} \quad Abhinav Gupta}

\date{Carnegie Mellon University}
\maketitle

\begin{abstract}
Self-supervised learning (SSL) aims to eliminate one of the major bottlenecks in representation learning - the need for human annotations. As a result, SSL holds the promise to learn representations from data in-the-wild, i.e., without the need for finite and static datasets. Instead, true SSL algorithms should be able to exploit the continuous stream of data being generated on the internet or by agents exploring their environments. But do traditional self-supervised learning approaches work in this setup? In this work, we investigate this question by conducting experiments on the \textit{continuous self-supervised learning problem}. 
While learning in the wild, we expect to see a continuous (infinite) non-IID data stream that follows a non-stationary distribution of visual concepts. The goal is to learn a representation that can be robust, adaptive yet not forgetful of concepts seen in the past. We show that a direct application of current methods to such continuous setup is 1) inefficient both computationally and in the amount of data required, 2) leads to inferior representations due to temporal correlations (non-IID data) in some sources of streaming data and 3) exhibits signs of catastrophic forgetting when trained on sources with non-stationary data distributions. We propose the use of replay buffers as an approach to alleviate the issues of inefficiency and temporal correlations. We further propose a novel method to enhance the replay buffer by maintaining the least redundant samples. Minimum redundancy (MinRed) buffers allow us to learn effective representations even in the most challenging streaming scenarios composed of sequential visual data obtained from a single embodied agent, and alleviates the problem of catastrophic forgetting when learning from data with non-stationary semantic distributions.
\end{abstract}

\section{Introduction}
\label{sec:intro}

\begin{figure}[t!]
    \centering
    \includegraphics[width=0.6\linewidth]{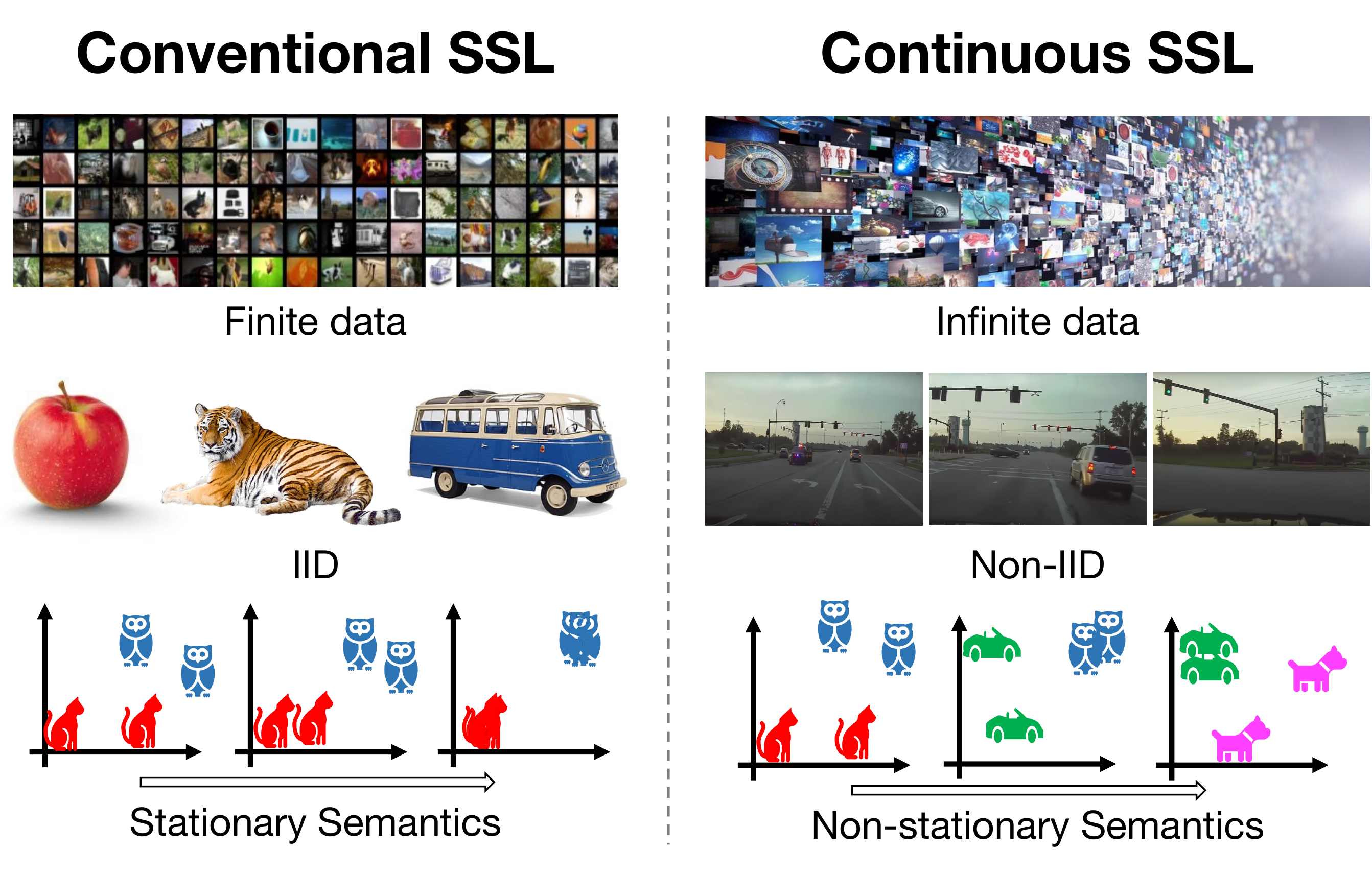}
    \caption{Conventional vs. Continuous Self-Supervised Learning. The conventional setup of fixed datasets for SSL violates key properties exhibited by data continuously gathered in-the-wild: infinite, non-IID and non-stationary semantics. Hence, for SSL methods that aim to be deployed in-the-wild, the conventional setup serves as a poor benchmark. In this work, we introduce the problem of continuous self-supervised learning to facilitate the evaluation of such methods and expose novel challenges.}
    \label{fig:teaser}
\end{figure}

We are witnessing yet another paradigm shift in the field of computer vision: from supervised to self-supervised learning (SSL). This shift promises to unleash the true potential of data, as we are no longer bound by the cost of manual labeling. Unsurprisingly, recent work has begun to scale current methods to extremely large datasets of up to 1 billion images~\cite{benchmarking,caron2019unsupervised,seer,swav,goyal2022vision} with the hope of learning better representations. In this paper, we pose the question: \textit{Are we ready to deploy SSL in-the-wild to harness the full potential of unlimited data?}

While SSL promises to exploit the infinite stream of data generated on the internet or by a robotic agent, current practices in SSL still rely on the traditional dataset setup. Images and videos are accumulated to create a training corpus, followed by optimization on hundreds of shuffled passes through the data. The primary reason for working with datasets is the need for reproducible benchmarks, but one question remains: is this traditional static learning setup right for benchmarking self-supervised learning? Does this setup accurately reflect the challenges of a self-supervised system deployed in the wild? We believe the answer is NO. 
For example, consider a self-supervised system attempting to learn representations of cars over the years from the web. Current setups only evaluate static learning and do not evaluate the ability to adapt representations to new car models (and not forget old ones). 
Another example is to consider a deployed robotic self-supervised learning agent that actively collects frames from its video feed. This data is heavily structured and correlated due to temporal coherence. However, existing SSL benchmarks do not reflect this challenge since they rely on datasets that can be randomly sampled to produce \textit{IID} samples.

In this paper, we move past dataset-driven SSL and investigate the efficacy of existing methods on the {\bf Continuous Self-Supervised Learning} problem. More specifically, we explore the challenges faced in two possible methods of deployment: (a) an internet-based SSL model which relies on continuously acquired images/videos; (b) an agent-based SSL system that learns directly from an agent's sensors. Both settings rely on a streaming data source that continuously generates new data, presenting three unique challenges that should be reflected when benchmarking SSL approaches (see \cref{fig:teaser}).

First, storing infinite amounts of data is not feasible and obtaining data in the wild often incurs a cost of time due to bandwidth or sensor speed limitations. As a result, epoch-based training is impossible, and a naive deployment of conventional SSL approaches, using each sample only once, would lead to inefficient learners, often waiting for data to be made available, while under-utilizing the data at its disposal. 
One solution is to rely on replay buffers to decouple data acquisition from the training pipeline. The first question we pose is how effective replay mechanism are at allowing representations to continue to improve while data is being collected?

Second, streaming data sources cannot be ``shuffled'' to create mini-batches of IID samples. Instead, the ordering of samples is dictated by the source itself. We show that this creates challenges for conventional representation learning approaches, as training data is not necessarily IID. Hence, we also pose the question of how to adapt existing SSL methods to learn robust representations under various non-IID conditions?

Third, real-world data is non-stationary. For example, a higher number of football-related images are seen during the world cup. Also, robots exploring indoor environments observe temporally clustered semantic distributions - a sequence of bedroom objects, followed by a sequence of kitchen objects, and so on. An intelligent lifelong learning system should be able to continuously learn new concepts without forgetting old ones from non-stationary data distributions. However, we show empirically that conventional contrastive learning approaches can overfit their representations to the current distribution, displaying signs of forgetting. We thus pose the question of how to design SSL methods that can learn under non-stationary conditions?

Overall, the main contributions of this work can be summarized as follows. We identify three critical challenges that arise in the continuous self-supervised learning setup, namely, training efficiency, robustness to non-IID data streams and learning under non-stationary semantic distributions. 
For each challenge, we construct a curated data stream that simulates this challenge and quantitatively demonstrate the shortcomings of existing SSL methods. We also propose initial solutions to these problems, with the goal of encouraging further research along these directions. We explore the idea of Buffered SSL, which involves augmenting existing approaches with a \textit{replay buffer} to improve training efficiency. Second, we propose a novel method to handle non-IID data streams by decorrelating stored samples. Finally, we show that \textit{decorrelated buffers} also prevent forgetting and improve continual learning under non-stationary data distributions.

\section{Related Work}
\label{sec:related_work}

Self-supervised visual representation learning is now a mature area of research, capable of producing models that even outperform fully supervised methods when transferred to a variety of downstream tasks~\cite{moco,simsiam,swav,byol}.
Despite forgoing the use of labeled data, these methods are still trained on fixed-size curated datasets originally developed for the supervised setting. This paper explores the various challenges of deploying self-supervised learning systems truly in-the-wild.

\begin{figure}[t!]
    \centering
    \includegraphics[width=0.95\linewidth]{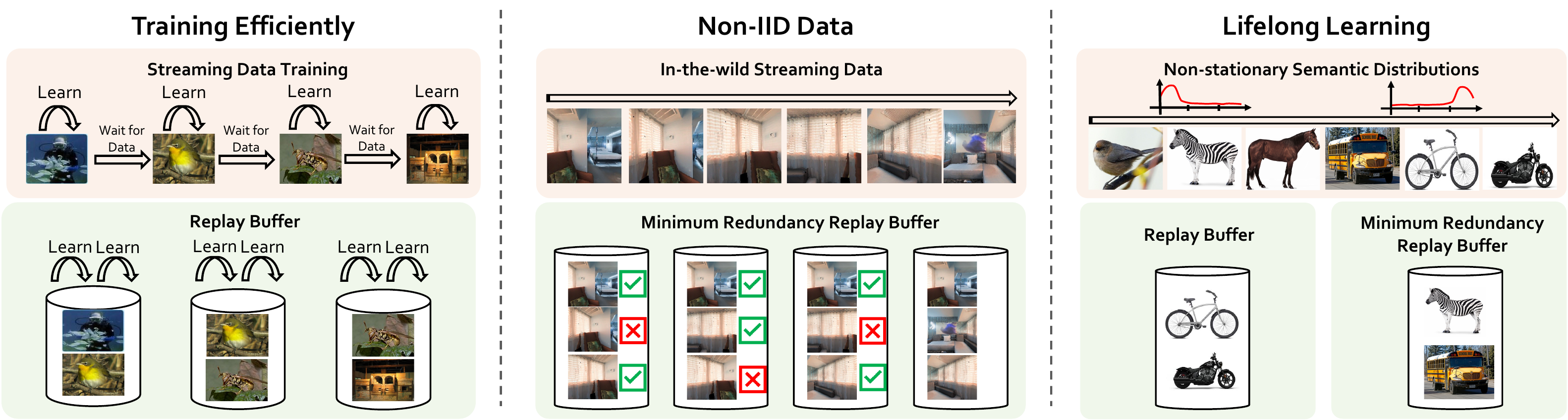}
    \caption{\textbf{Overview}: We investigate the problem of continuous self-supervised learning, exposing three challenges faced by SSL methods deployed in-the-wild. First, the infinite nature of data streams implies that samples cannot be repeated. We show that augmenting an existing SSL method~\cite{simsiam} with replay buffers can significantly alleviate this issue. Second, data gathered continuously in-the-wild is often temporally correlated, violating the IID assumption of optimization algorithms. We show that enhancing replay buffers to maintain minimally redundant samples (MinRed), we can generate data that are less correlated. Finally, semantic distributions of data gathered in-the-wild are non-stationary. This poses the challenge of ``forgetting" concepts seen in past distributions. We show that MinRed buffers can also alleviate the issue of ``forgetting" by collecting unique samples from various semantic groups. }
    \label{fig:overview}
\end{figure}

\textbf{Self-supervised learning} has a long history in computer vision~\cite{mobahi2009deep,masci2011stacked,salakhutdinov2009deep,de1994learning,le2013building,deep_cluster} aiming to learn representations of visual data by solving tasks that can be defined without human annotations. A breadth of methodologies has been proposed from generative models such as denoising auto-encoders~\cite{vincent2008extracting}, sparse coding \cite{lee2007efficient,olshausen1996emergence,olshausen2000sparse}, inpainting~\cite{inpainting} and colorization~\cite{colorization,larsson2016learning,deshpande2015learning}, to methods that learn representations predictive of spatial context~\cite{context_pred,jigsaw,rotation}, temporal context~\cite{misra2016shuffle,pathak2017learning,wang2015unsupervised,odd1out,qian2021spatiotemporal,pace}, or concurrent modalities like audio~\cite{de1994learning,owens2018audio,morgado2021audio,arandjelovic2017look}, text~\cite{clip,gomez2017self,desai2021virtex} or speech~\cite{howto,miech2020end}.

One successful approach is to learn transformation invariant representations~\cite{hadsell2006dimensionality,ranzato2007unsupervised,exemplar,wu2018unsupervised,cpc,simclr,pirl,moco}.
Prior work has developed improved image augmentations~\cite{simclr,pirl}, backbone models~\cite{dino,seer}, stable (slow-moving) learning targets~\cite{moco,mocov3,swav}, and transformation invariant loss functions~\cite{cpc,byol,zbontar_barlowtwins,dino,simsiam}. As a result, SSL has produced impressive models that improve state-of-the-art on a diverse set of downstream tasks like recognition~\cite{swav,dino}, detection~\cite{moco} and video object segmentation~\cite{dino}. 

Given its success, a few attempts have been made to scale SSL to large uncurated datasets, such as YFCC-100M~\cite{benchmarking,caron2019unsupervised} and Instagram-1B~\cite{swav,seer}.
Goyal \textit{et al.}~\cite{benchmarking} showed that tasks such as colorization~\cite{colorization}, context prediction~\cite{jigsaw} and rotation~\cite{rotation} have diminishing returns on large datasets, due to the low complexity of the task, and argued for the development of more complex tasks. Transformation invariance objectives, coupled with heavy data augmentations, have increased the task's complexity substantially. As a result, recent attempts of scaling up augmentation invariance~\cite{swav,seer,goyal2022vision} have seen some performance gains.
However, we argue that these methods are still not ready to be deployed truly in-the-wild.
Beyond the difficulties of training on uncurated data, already studied in prior work~\cite{swav,seer}, training on fixed datasets ignores important challenges of streaming data, such as the non-iid nature of streaming sources, data acquisition costs, and model saturation due to its fixed capacity.

\textbf{Continual and lifelong learning:}
The ability to continuously learn new concepts or tasks over time is often referred to as lifelong learning~\cite{thrun1995lifelong} or never-ending learning~\cite{chen2013neil,mitchell2018never}. Lifelong learning has traditionally been studied in supervised and reinforcement learning settings. In both cases, the model is expected to learn from a set of distinct tasks presented sequentially, without forgetting previous ones~\cite{lwf,kirkpatrick2017overcoming,rannen2017encoder,zenke_continual,titsias2019functional}. 
However, these works usually assume access to full supervision in the form of class labels or external rewards, not available in the streaming setup. 

Techniques developed for supervised continual learning are nevertheless useful in the Continuous SSL problem. Rehearsal techniques~\cite{rebuffi2017icarl,aljundi2019gradient,rolnick2019experience,shin2017continual,dark_replay} store and replay a small set of training samples from previous tasks to avoid forgetting previously learned skills or concepts. While there is no notion of well-defined tasks in Continuous SSL, we show that replay buffers help improve training efficiency. We also propose replay buffers that minimize the redundancy of stored memories to decorrelate highly correlated streaming sources.
Beyond rehearsal techniques, expandable models~\cite{rusu2016progressive,yoon2017lifelong} have also been used to reduce catastrophic forgetting in supervised continual learning. This is often accomplished either by progressively growing the model each time a new task is added~\cite{rusu2016progressive,yoon2017lifelong,li2019learn}, or maintaining a common backbone model which is adapted to each task separately using small task-specific adaptation blocks~\cite{rebuffi2017icarl,mallya2018piggyback,morgado2019nettailor}. 
The lack of well-defined tasks in streaming SSL makes lifelong learning more challenging, as it needs to learn from data distributions that may shift over time. 

\noindent {\bf Lifelong Generative Models:} None of the existing literature has investigated how discriminative self-supervised representation learning methods perform in the full continuous learning setup (streaming, non-IID and non-stationary data). However, recent works~\cite{ye2020learning,ramapuram2020lifelong,achille2018life,rao2019continual} have attempted to address a sub-problem of ours, \ie, learning self-supervised representations using generative models in a continual learning setting where the domain of data exhibits significant shifts during training. 
These works present approaches to locate domain shifts in order to avoid the problem of catastrophic forgetting. These techniques are made possible by the fact that training data is constructed by collecting samples from images in significantly different datasets - for example, \cite{ye2020learning} uses Celeb-A\cite{liu2015deep} faces followed by 3D-Chair\cite{aubry2014seeing} images). In contrast, we consider a more realistic setting of ImageNet images with a smoothly changing distribution of classes.
Furthermore, as highlighted above, these works do not address other critical challenges of deploying SSL in-the-wild, as they are limited to epoch-based optimization, do not consider non-curated and/or high correlated streaming sources, data efficiency, or the issue of early convergence. 
\section{Problem Setup and Challenges}
\label{sec:method}

The goal of this work is to investigate the efficacy of self-supervised representation learning on a continuous source of streaming data generated in the real world, which we refer to as the \textit{continuous self-supervised learning problem}. First, we describe the distinction between conventional training and the continuous self-supervised learning setup. We then discuss the various unique challenges that appear in the continuous case.

\vspace{-5pt}
\subsection{Streaming vs Conventional Self-Supervised Learning}\vspace{-5pt}

\noindent Existing self-supervised learning methods rely on fixed-size datasets. These datasets $\Dcal=\{\x_1,\ldots,\x_N\}$ are finite (\ie, $N<<\infty$), immutable (\ie, $\Dcal$ does not change) and readily available (\ie, all its samples $\x_i$ can be easily accessed at all times). Due to these properties, samples can be indexed, shuffled, and accessed at any point in training. Conventional SSL takes advantage of these possibilities by iterating over the datasets multiple times (epochs). 

In contrast, Continuous SSL relies on a \textit{streaming source} $\Scal$, defined as a time-series of unlabeled sensory data $\Scal=(\x_1,\x_2,\ldots,\x_T)$, potentially of infinite length $T\xrightarrow{}\infty$. At any given moment in time $t$, fetching data from a streaming source $\Scal$ yields the current sample $\x_t$. Future samples $\{\x_\tau \forall\tau>t\}$ are not accessible at time $t$, and past samples $\{\x_\tau \forall\tau<t\}$ are only accessible if stored when fetched. 

In the Continuous SSL setup, one important parameter is the ratio between the data loading time $t_\text{data}$ and the time taken to perform one optimization step $t_\text{opt}$. In most deployment setups $t_\text{data} > t_\text{opt}$, due to slower data transfer speed or low sensor frame rates. Therefore, even with parallelization, optimization algorithms can wait idle for $t_\text{idle} = t_\text{data}-t_\text{opt}$. Therefore, SSL methods developed for the continuous setup should be able to efficiently and continually build better representations, while training on samples obtained from a streaming source. 

\vspace{-5pt}
\subsection{Why Continuous SSL? Does scaling the number of unique images help representation learning?}
\vspace{-5pt}

\noindent To understand the effect of increasing the scale of training data (potentially to infinite), we indexed all Creative Commons images uploaded to the photo-sharing website Flickr.com between 2008 and 2021. We then used this index to create datasets of varying sizes, and train visual representations through self-supervision over multiple epochs in the Conventional SSL setup.

We adopt SimSiam~\cite{simsiam} as a prototypical example of contrastive learning methods, which have been shown to be effective for Conventional SSL. SimSiam learns representations by optimizing the augmentation invariance loss
\begin{equation}
    \Lcal(x_1, x_2) = - \sg(\z_1)^T g(\z_2) - \sg(\z_2)^T g(\z_1)
\end{equation}
\begin{wrapfigure}[14]{r}{0.5\textwidth}
    \centering
    \vspace{-10pt}
    \includegraphics[width=\linewidth]{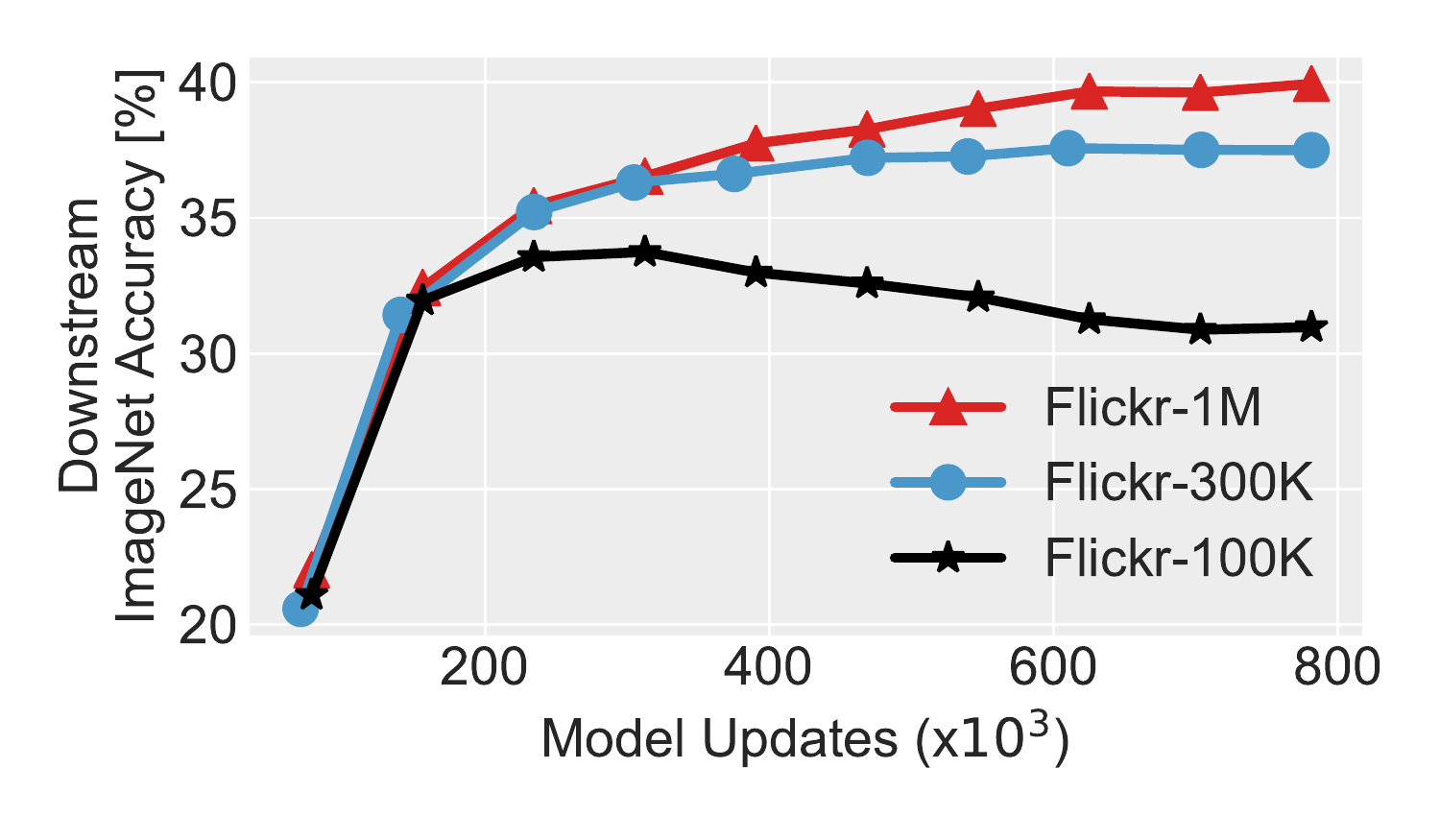}
    {\footnotesize
    \caption{ImageNet downstream accuracy of a SimSiam model trained on datasets of different sizes with a ResNet-18 backbone.}}
    \label{fig:db_size_exp}
\end{wrapfigure}
where $x_1$ and $x_2$ are two random transformations of an image $x$, $\z_i=f(x_i)$ is the model output representations, $\sg(\cdot)$ the stop gradient and $g(\cdot)$ a prediction head. Refer to \cite{simsiam} for full details.
\cref{fig:db_size_exp} shows the linear classification accuracy on ImageNet for models trained on different datasets as a function of the number of model updates. Unsurprisingly, training with more diverse data leads to better representations. 
This highlights the benefits of scaling \textit{unique} images, which Continuous SSL will take to the extreme.

\vspace{-5pt}
\subsection{Challenges of Continuous SSL}

Learning representations in the Continuous SSL setup poses novel challenges that Conventional SSL methods do not face.

\vspace{-6pt}\paragraph{\textbf{- Epochs vs One Pass}} Streaming sources do not allow revisiting samples obtained in the past unless they were stored. Since storing the full stream is infeasible due to the potentially infinite length, Continuous SSL methods are required to learn representations in ``one pass" over the samples.

\vspace{-6pt}\paragraph{\textbf{- Sampling Efficiency}} Sampling data from streaming sources in the real world can be inefficient due to sensor frame rates or bandwidth limitations. This significantly increases the time taken to learn representations as optimization algorithms may have to wait idly while waiting for data.

\vspace{-6pt}\paragraph{\textbf{- Correlated Samples}}
Many streaming sources in the wild exhibit temporal coherence.
For example, consecutive frames from online videos or from a robot exploring its environment display minimal changes. Such correlations break the \textit{IID} assumption on which conventional optimization algorithms rely.

\vspace{-6pt}\paragraph{\textbf{- Lifelong learning}}
Access to infinite streams of data provides us the opportunity to continuously improve visual representations. However, the non-stationary nature of data streams in the wild cause conventional SSL methods to quickly forget features that are no longer relevant for the current distribution. This poses another challenge: as we continuously acquire new data, how can Continuous SSL methods integrate new concepts in their representations without forgetting previously learned ones?

\vspace{6pt}
While all these challenges co-exist in the wild, evaluating current SSL methods directly would prevent us from analyzing each one comprehensively and in isolation. Instead, we disentangled each challenge by designing a set of data streams that highlight each problem separately, and assess its effect on existing SSL methods. This helps us building a thorough characterization of each challenge and inform us on how to tackle them. We believe a disentangled analysis will help the community build intuitions about the impact of each challenge on continuous SSL as a whole. Section \cref{sec:efficiency} introduces the challenge of one pass training and computational efficiency. Section~\cref{sec:correlated} introduces the non-iid data setup, and Section~\cref{sec:lifelong} analyses the lifelong learning setting.

\section{Efficient Training}
\label{sec:efficiency}
Computational and data efficiency are two challenges that currently prevent SSL from being deployed on continuous data streams in-the-wild. For most practical applications, $t_\text{data} : t_\text{optim}$ might be high, so SSL methods should use idle time to improve the models. Second, fetching new samples can still be costly. For example, exploration robots often run on batteries, and web crawlers are limited by network bandwidths. Trivially deploying current SSL methods to the streaming setup would discard each batch of data after being used once. However, current deep learning optimization practices show that iterating over the same samples over multiple epochs helps learn better representations. For example, supervised learning on ImageNet~\cite{alexnet,resnet} iterates over the dataset 100 times, and SSL approaches~\cite{simclr} have been shown to keep improving even after seeing each sample 800 times. Therefore, we would like to answer the question of how to improve data efficiency while still following the streaming setting.

\subsection{Buffered Self-Supervised Learning}
\label{sec:buffer-ssl}
\begin{figure}[t!]
    \centering
    \begin{subfigure}{0.7\linewidth}
        \centering
        \includegraphics[width=0.9\linewidth]{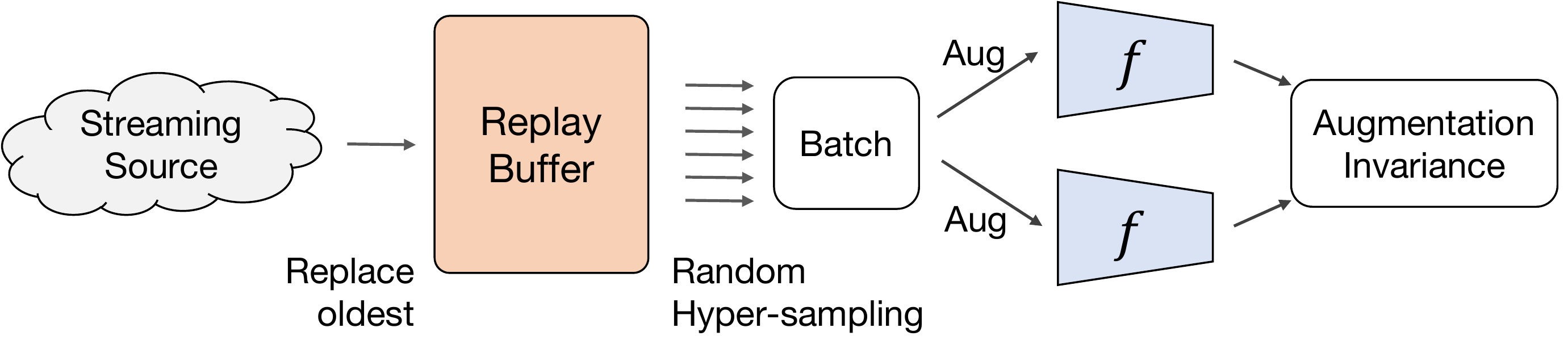}
        \caption{Overview of Buffered SSL}
        \label{fig:buff-ssl-overview}
    \end{subfigure}\\\vspace{5pt}
    \begin{subfigure}{0.7\linewidth}
        \centering
        \includegraphics[width=0.9\linewidth]{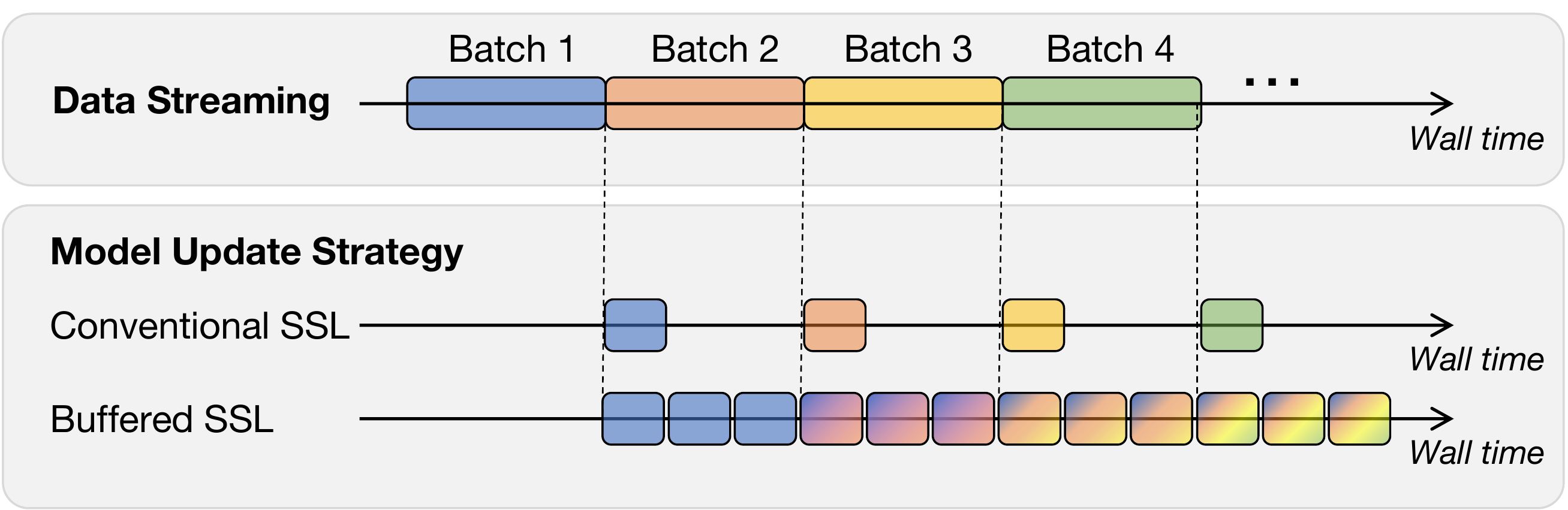}
        \caption{Optimization under limited streaming bandwidth.}
        \label{fig:bandwidth-vis}
    \end{subfigure}
    \caption{\textbf{Buffered Self-Supervised Learning.} Buffered SSL introduces a replay buffer, which allows the model to continuously train even under limited bandwidth settings.}
    \label{fig:buff-ssl}
\end{figure}

We present a simple solution to the challenges above. The key idea is to maintain a fixed-size \textit{replay buffer} that stores a small number of recent samples. This idea is inspired by experience replay~\cite{lin1992reinforcement} commonly used in reinforcement learning~\cite{mnih2015human,schaul2016prioritized,andrychowicz2017hindsight} and supervised continual learning~\cite{hsu2018re,rolnick2019experience}. As shown in \cref{fig:buff-ssl-overview}, the replay buffer decouples the streaming source from the training pipeline. 
The streaming data can be added to the replay buffer when available, replacing the oldest samples (\textit{i.e.} first-in-first-out (FIFO) update rule). Simultaneously, mini-batches of training data can be generated at any time by randomly sampling from the buffer. 
As shown in~\cref{fig:bandwidth-vis}, replay buffers allow us to continue training during the otherwise idle wait time $t_\text{idle}$. Replay buffers also allow us to reuse samples by sampling them multiple times, hence reducing the total data cost. We refer to this approach as \textit{Buffered Self-Supervised Learning}.

\begin{SCfigure}[][t!]
    \includegraphics[width=0.44\linewidth,trim={0 25 0 15},clip]{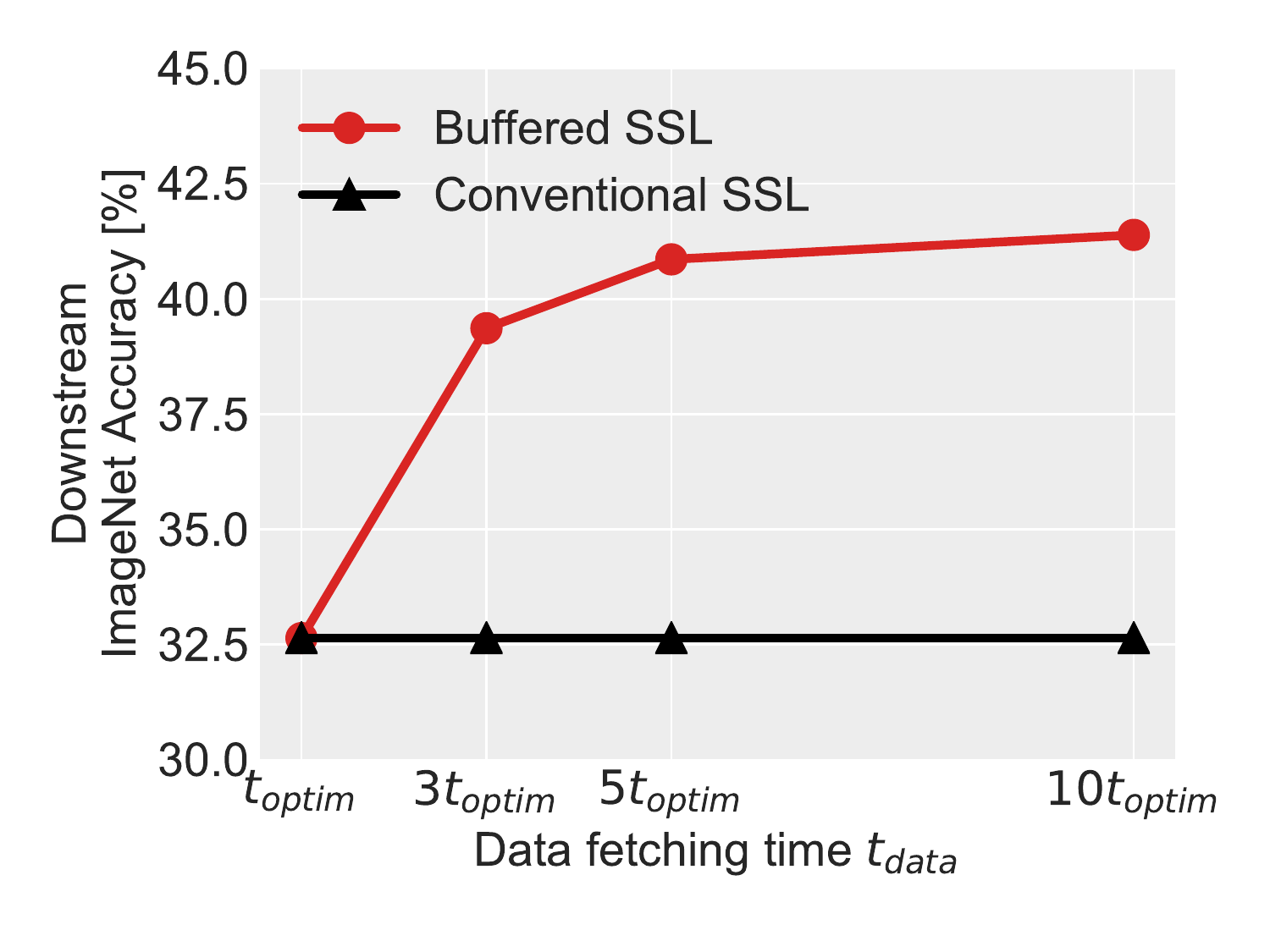}
    \caption{\textbf{Streaming SSL with limited bandwidth.} Comparison of buffered and non-buffered approaches for various limited bandwidth settings.  $t_\text{data} : t_\text{optim}$ denotes the ratio of data acquisition time to the optimization time. Buffered SSL can take advantage of the idle time to effectively improve the learned representations instead of waiting idly for new data.}
    \label{fig:bandwidth}
\end{SCfigure}

\subsection{Single-pass training experiments}
We study the effectiveness of replay buffers when training with a single pass of the data. We trained ResNet-18 SimSiam models with and without replay buffers, with various amounts of idle time $t_\text{idle}=t_\text{data}-t_\text{optim}$. 
All models were trained using the first 20 million images in our Flickr index as the streaming source.

\begin{table}[t]
    \centering
    \caption{\textbf{Data Efficiency:} Augmenting SSL methods with replay buffers can improve efficiency allowing us to train on data streams with one pass. We show that Buffered SSL methods outperform the Conventional SSL methods and achieve performances close to training for multiple epochs.}
    \label{tab:data_efficiency_exp}
    \resizebox{0.7\linewidth}{!}{
    \begin{tabular}{@{\extracolsep{\fill}}lccccc@{}}
    \toprule
    & \thead{Epochs} & \thead{Hyper\\Sampling} & \thead{Memory\\Size} &   \thead{ImageNet\\Top1 Acc}  & \thead{iNaturalist\\Top1 Acc} \\ \midrule

    \rg\multicolumn{6}{l}{\textit{Training DB: Flickr 20M}}\\
    Conventional SSL & 1 & -   & - & 32.3 & 9.8 \\
    Buffered SSL  & 1  & 10 & 16K  & 41.4 & 16.7 \\ 
    Buffered SSL  & 1  & 10 & 64K  & 41.8 & 17.3 \\ 
    Buffered SSL  & 1  & 10 & 256K & 41.5 & 17.5 \\ 
    \arrayrulecolor{lightgray}\hline\arrayrulecolor{black}
    Epoch-based SSL${}^*$ & 10 & -  & -    & 41.9 & 17.5 \\
    \midrule
    \rg\multicolumn{6}{l}{\textit{Training DB: Flickr 5M}}\\
    Conventional SSL & 1 & - & -  & 14.5 & 2.8 \\
    Buffered SSL  & 1  & 40 & 16K  & 39.9 & 16.1 \\
    Buffered SSL  & 1  & 40 & 64K  & 41.0 & 17.1 \\
    Buffered SSL  & 1  & 40 & 256K & 41.5 & 17.3 \\
    \arrayrulecolor{lightgray}\hline\arrayrulecolor{black}
    Epoch-based SSL${}^*$ & 40 & -  & -  & 41.8 & 17.0 \\
    \midrule
    \rg\multicolumn{6}{l}{\textit{Training DB: Flickr 1M}}\\
    Conventional SSL & 1 & -   & -    & 8.0 & 1.5 \\
    Buffered SSL  & 1   & 200 & 16K  & 30.5 & 9.5  \\
    Buffered SSL  & 1   & 200 & 64K  & 36.4 & 14.3 \\
    Buffered SSL  & 1   & 200 & 256K & 38.8 & 15.5 \\
    \arrayrulecolor{lightgray}\hline\arrayrulecolor{black}
    Epoch-based SSL${}^*$ & 200 & -   & -    & 41.7 & 17.3 \\
    \bottomrule
    \multicolumn{6}{l}{${}^*$\textit{Epoch-based SSL violates the streaming setting (reference only).}}
    \end{tabular}
    }\vspace*{-0.1in}
\end{table}

\cref{fig:bandwidth} shows the ImageNet linear classification performance for increasing $t_\text{data}$.
By maintaining a small replay buffer (containing only the most recent 64k images), Buffered SSL was able to make good use of the idle time and improve representations significantly (41.4\% accuracy on ImageNet) over the bottlenecked Conventional SSL approach (32.5\% ImageNet accuracy).
Replay buffers also improve data efficiency in the Continuous SSL setup, as each sample can be reused multiple times. 
Data usage is proportional to the hyper-sampling rate $K$, defined as the ratio between the number of mini-batches generated for training and acquired from the streaming source.

To understand the limits of hyper-sampling, we trained a ResNet-18 SimSiam model with a replay buffer for a fixed amount of updates (780\,000 iterations).
\cref{tab:data_efficiency_exp} shows a comparison of Buffered SSL at varying hyper-sampling rates $K$, to Conventional SSL trained on the same amount of data, and Epoch-based SSL methods trained for $K$ epochs. Epoch-based SSL and Buffered SSL are optimized with the same number of updates, but the former violates the streaming setup.
Despite being required to train on a single pass of the data, Buffered SSL with a hyper-sampling rate of $K=10$ achieved similar performance to epoch-based training, even for buffers as small as 64K images (0.3\% of the 20M unique images seen).
\cref{tab:data_efficiency_exp} also shows that, as hyper-sampling rates increase, the size of the replay buffer becomes critical. 
For example, for $K=200$, Buffered SSL still improves significantly over Conventional SSL on the same amount of data, regardless of buffer size. However, better representations are learned as the buffer size increases.
Since, in high hyper-sampling regimes, the buffer is updated slowly with new images from the streaming source, increasing the buffer size prevents the model from quickly overfitting to the samples in the buffer.

\section{Correlated Data Sources}
\label{sec:correlated}

\begin{table}[t!]
    \centering
    \caption{\textbf{Visually Correlated SSL:} Linear classification performance of buffered and unbuffered SimSiam representations trained on data sources with high temporal coherence. MinRed buffers learns better representations by decorrelating the data.}
    \label{tab:vis_corr}
    \resizebox{0.7\linewidth}{!}{
    \begin{tabular}{@{\extracolsep{\fill}}lccccc@{}}
    \toprule
    & \thead{Epochs} & \thead{Hyper\\Sampling} & \thead{Memory\\Size} & \thead{ImageNet\\Top1 Acc} & \thead{iNaturalist\\Top1 Acc} \\ 
    \midrule
    \rg\multicolumn{6}{l}{\textit{Streaming source: Kinetics $(N_\text{seq}\!=\!16)$}} \\
    Conventional SSL        & 5 & -  & -    & 17.7 & 3.0 \\
    Buffered SSL            & 1  & 5 & 64K  & 25.9 & 8.4 \\ 
    Buffered SSL (MinRed)   & 1  & 5 & 64K  & 26.2 & 7.9 \\ 
    \arrayrulecolor{lightgray}\hline\arrayrulecolor{black}
    Decorrelated source${}^*$ & 5  & - & -  & 25.9 & 7.9 \\
    \midrule
    \rg\multicolumn{6}{l}{\textit{Streaming source: Kinetics $(N_\text{seq}\!=\!64)$}} \\
    Conventional SSL        & 5 & -  & -    & 7.6 & 0.8 \\
    Buffered SSL            & 1  & 5 & 64K  & 11.7 & 1.4 \\ 
    Buffered SSL (MinRed)   & 1  & 5 & 64K  & 31.2 & 9.9 \\ 
    \arrayrulecolor{lightgray}\hline\arrayrulecolor{black}
    Decorrelated source${}^*$ & 5  & - & -  & 30.7 & 9.9 \\
    \midrule
    \rg\multicolumn{6}{l}{\textit{Streaming source: Krishna CAM}} \\
    Conventional SSL        & 5 & -  & -    & 0.4  & 0.03 \\
    Buffered SSL            & 1  & 5 & 16K  & 0.5  & 0.05 \\ 
    Buffered SSL (MinRed)   & 1  & 5 & 16K  & 15.2 & 3.43 \\ 
    Buffered SSL            & 1  & 5 & 64K  & 1.7  & 0.07 \\ 
    Buffered SSL (MinRed)   & 1  & 5 & 64K  & 17.9 & 5.91 \\
    \arrayrulecolor{lightgray}\hline\arrayrulecolor{black}
    Decorrelated source${}^*$ & 5  & - & -  & 19.2 & 6.94 \\
    \bottomrule
    \multicolumn{6}{l}{${}^*$\textit{Decorrelated sources violate the streaming setting (reference only).}}
    \end{tabular}
    }
\end{table}

Visual data obtained in-the-wild is often correlated and non-\textit{IID}. For example, video feed from a self-driving car collects very similar consecutive frames. This is in stark contrast to the data used in Conventional SSL methods. For example, the ImageNet dataset allows sampling images from a collection of 1000 uniformly distributed object classes. Even methods trained on larger datasets like Instagram-1B~\cite{benchmarking,seer} are less likely to encounter heavily correlated samples in the mini-batches. 
However, the constant flow of data in the Continuous SSL setup generally violates these assumptions even in the static image setup (images uploaded near events are likely to be highly correlated).

Let $(x_i: i\in \mathcal{D})$ be a sequence of samples. When $x_i$ is generated by randomly sampling from a large dataset, samples are close to IID. Hence, the probability $p_c$ that two samples $x_i$ and $x_j$ are highly correlated is low, $p_c \approx 0$. Correlated samples may indicate images that are visually very similar or visually dissimilar but depict similar semantic content.
However, in the Continuous SSL setup, the IID assumption is generally violated, leading to $p_c>>0$.
Under the assumption that consecutive samples in a continuous stream of data have the same correlation probability $p_c$, the likelihood of a random pair in a batch $(x_i,\ldots,x_{i+b})$ of size $b$ being correlated (\textit{correlation likelihood}) is large, and given by
{\small
\begin{equation}
    \label{eq:l_corr}
    \mathcal{L}_\text{Seq} = P_c(b, p_c) = \frac{2}{b(b-1)}\sum_{i=1}^{b-1}\sum_{j=i+1}^b p_c^{j-1} 
    = \frac{2p_c}{b(b-1)} \Big( \frac{p_c^b-1}{(1-p_c)^2}+b\frac{p_c}{1-p_c}\Big).
\end{equation}
}

Introducing a replay buffer of size $B>>b$, as proposed in \cref{sec:buffer-ssl}, lowers the correlation likelihood to $\mathcal{L}_\text{FIFO} = P_c(B,p_c)\approx\frac{b}{B}\mathcal{L}_\text{Seq} < P_c(b, p_c)$\footnote{Approximation holds for large values of $B$ and $b$, and $p_c\neq1$.}, and enables more effective representation learning. 

\begin{figure}[t!]
    \centering
    \begin{subfigure}{0.49\linewidth}
        \includegraphics[width=\linewidth,trim={0 10 0 10},clip]{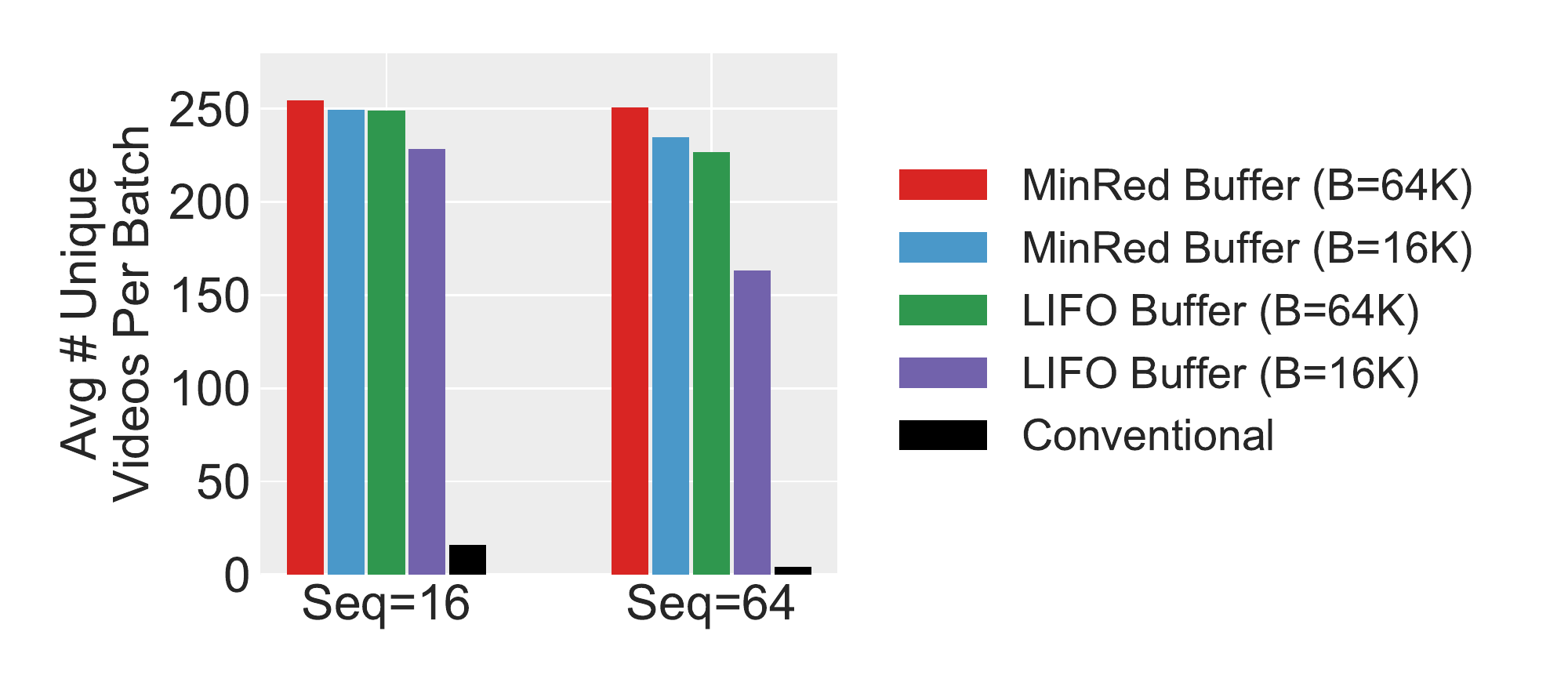}
        \caption{Kinetics}
        \label{fig:correlation-kinetics}
    \end{subfigure} 
    \begin{subfigure}{0.42\linewidth}
        \includegraphics[width=\linewidth,trim={0 10 0 10},clip]{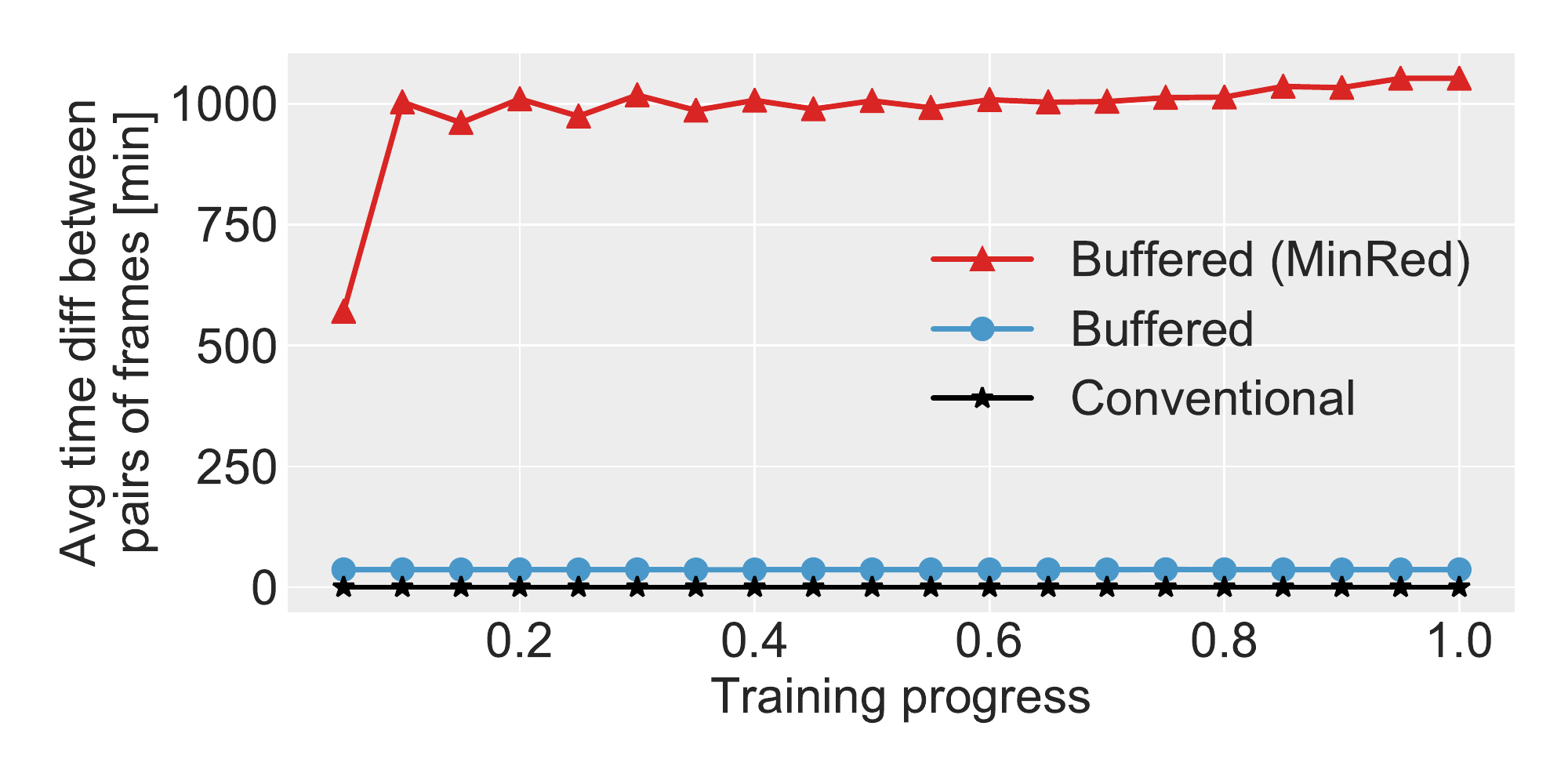}
        \caption{Krishna CAM}
        \label{fig:correlation-krishna}
    \end{subfigure}
    \caption{Estimate of within batch correlation while training w/ and w/o replay buffers.}
    \label{fig:correlation_pc}
\end{figure}

\subsection{Minimum Redundancy Replay Buffer}
\label{sec:minred}
While replay buffers are able to reduce the correlation likelihood, prohibitively large replay buffers ($B>>b$) are required to significantly lower $\mathcal{L}_\text{FIFO}$ in heavily correlated setups ($p_c\approx1$). In order to overcome this, we propose a modified replay buffer to only retain de-correlated samples, thereby actively reducing $p_c$. We call this the Minimum Redundancy Replay Buffer (MinRed).

To accomplish this, we rely on the learned embedding space to identify redundant samples. Consider a replay buffer $\Bcal$ with a maximum capacity of $B$, already containing $B$ samples with representation $\bar{\z}_i$. To add a new sample $x$ to $\Bcal$, we rely on the cosine distance between all pairs of samples to discard the most redundant:
\begin{equation}
    \Bcal \leftarrow \Bcal \backslash i^* \cup \{x\}
    \quad \mbox{where} \quad i^* =\arg\min_{i\in\Bcal} \min_{j\in\Bcal} d_{\mathtt{cos}}(\bar{\z}_i, \bar{\z}_j).
\end{equation}
In other words, we discard the sample with minimum distance to its nearest neighbor. To represent instances, we track the features $\bar{\z}_i$ of all samples in the buffer using a moving average $\bar{\z}_i\!=\!\alpha \bar{\z}_i + (1-\alpha) \z_i$,
where $\z_i\!\!=\!\!f(\x_i)$ is the current feature of the $i^{th}$ sample, and $\alpha$ the moving average coefficient. 
Since redundant samples are dropped from the buffer, the probability $p_c$ of two consecutive samples in the buffer being correlated decreases. If this probability decreases from $p_c$ to $\eta p_c$ where $\eta<<1$, the correlation likelihood is lowered to $\mathcal{L_\text{MinRed}}\!=\!P_c(B,\eta p_c)<P_c(B,p_c)$, which facilitates representation learning.

\subsection{Experiments with non-IID data streams}

We assess the performance of SSL methods on two data streams with heavy temporal coherence. The first data stream is created by concatenating samples from videos in the Kinetics dataset~\cite{carreira2017quo}. From each video, we sample $N_\text{seq}$ frames at random and add them sequentially to the data stream. The second training stream is taken as consecutive frames from the KrishnaCAM dataset\footnote{Concatenated videos are looped over 10 times to create a large stream.}~\cite{singh2016krishnacam} which records ego-centric videos spanning nine months of the life of a computer vision graduate student. On each stream, we train the baseline SimSiam (Conventional SSL), SimSiam augmented with replay buffers (Buffered SSL) and SimSiam augmented with MinRed buffers (Buffered SSL (MinRed)). We evaluate these representations by training a linear classifier on the ImageNet~\cite{imagenet} and iNaturalist~\cite{inaturalist} datasets. Results are shown in Table~\ref{tab:vis_corr}. We observe that the correlated nature of the data heavily disrupts training of the conventional models. While the regular replay buffers alleviate this issue to some extent, learned representations still suffer when trained on heavy correlated data streams (as in Kinetics $N_\text{seq}\!=\!64$ and KrishaCAM). Finally, the proposed MinRed buffers demonstrate significant gains in these setups. Models trained with MinRed buffers are generally very close to the ``oracle" setting of training from completely decorrelated streams of data (\textit{i.e.} randomly sampling from the collection of all frames from all videos, and thus violating the streaming assumption).

\vspace{5pt}\noindent
\textbf{Correlation of training samples:} One of the benefits of Buffered SSL is the ability to generate training samples with low correlation likelihood and thus closer to \textit{IID}. We analyzed the contents of the replay buffer over the duration of training to track the correlation likelihood (see \cref{fig:correlation_pc}). We confirmed that the contents of MinRed replay buffers are significantly less correlated than FIFO buffers. In KrishnaCAM, MinRed buffers tend to maintain memories of past unique frames for longer periods of time. In Kinetics, MinRed buffers also yield training mini-batches with frames from a larger number of unique videos.

\section{Lifelong Self-Supervised Learning}
\label{sec:lifelong}

As we explore the world, we come across different distributions of object classes, some previously seen and some unseen. For example, we see furniture and appliances every day. But we also encounter novel concepts like zebras when we visit a zoo. This suggests that the distribution of semantic classes is often correlated in time with occasional changes in distribution. However, Conventional SSL methods learn from a limited vocabulary of concepts that is repeatedly seen thousands of times (often uniformly). This provides a simplification of the learning setup that does not reflect the non-stationary nature of concepts in-the-wild. 

\subsection{A non-stationary data stream to benchmark SSL}
To evaluate deployable SSL methods, we must use benchmarks that simulate the non-stationary semantic distributions we encounter in-the-wild. Inspired by supervised continual learning~\cite{lwf,kirkpatrick2017overcoming}, we introduce a setup with smooth shifting semantic distributions. Partitions will be made publically available.
\begin{figure}[t!]
    \centering
    \begin{subfigure}{0.62\linewidth}
        \includegraphics[width=\linewidth,trim={0 10 0 10},clip]{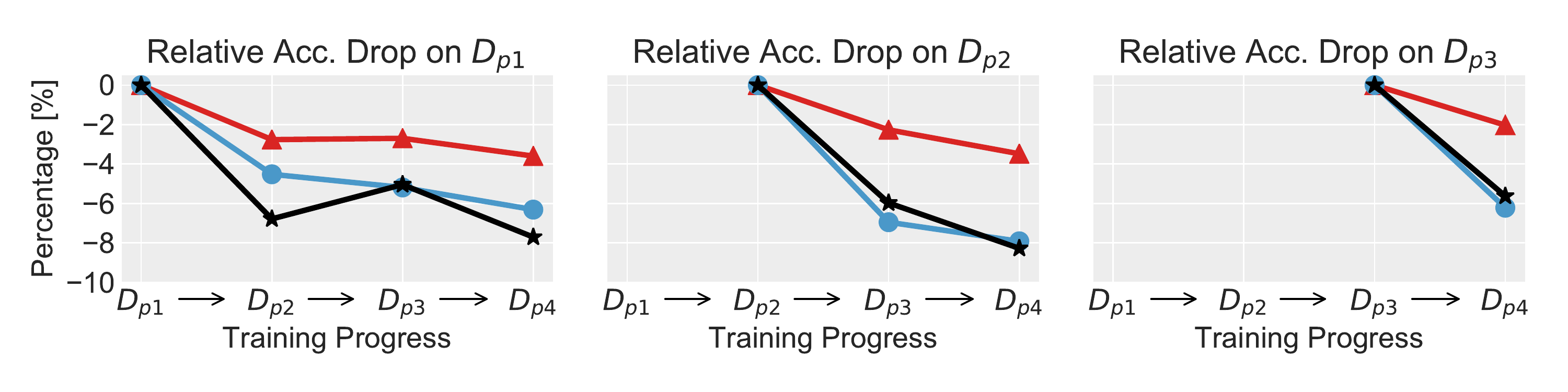}
        \caption{\textit{Forgetting}}
        \label{fig:continual-forgetting}
    \end{subfigure}\hfill
    \begin{subfigure}{0.36\linewidth}
        \includegraphics[width=\linewidth,trim={0 10 0 10},clip]{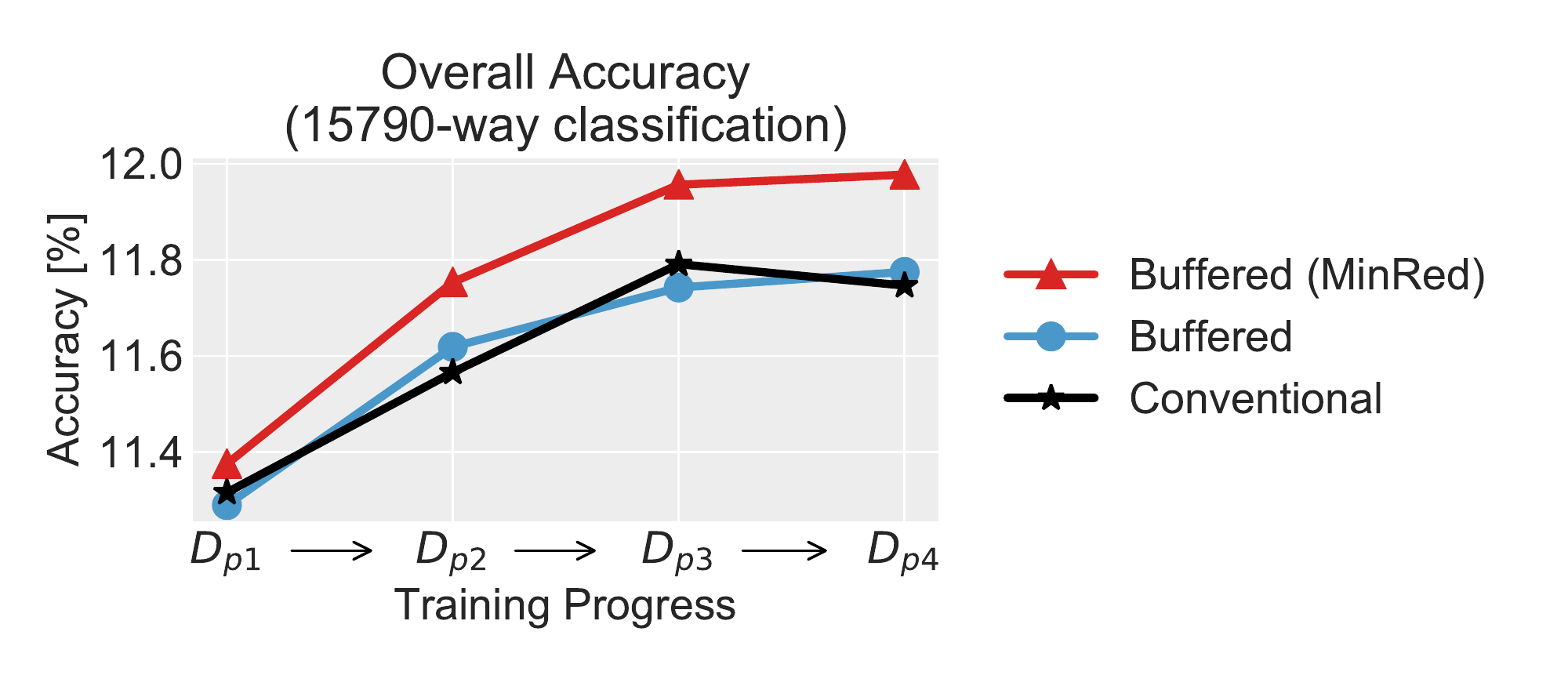}
        \caption{\textit{Generalization}}
        \label{fig:continual-accuracy}
    \end{subfigure}
    \caption{Continual unsupervised representation learning on full ImageNet (14M images). The dataset is partitioned in 4 separate tasks which are seen in a  sequence $D_{p1}$ $\rightarrow$ $D_{p2}$ $\rightarrow$ $D_{p3}$ $\rightarrow$ $D_{p4}$. Forgetting \ref{fig:continual-forgetting} is measured by computing the relative accuracy drop on each task after training on data of the task itself. 
    Minimum redundancy buffers naturally retain instances from previous tasks, thus mitigating the catastrophic forgetting observed with conventional SSL and regular replay buffers. Generalization \ref{fig:continual-accuracy} is measured as the overall accuracy across all 15790 full ImageNet classes. By ensuring that images from past class distributions are not forgotten, minimum redundancy buffers can learn better representations overall. All results are averaged over 3 different sequences $p_i$.}
    \label{fig:bandwith}
\end{figure}

First, we create four datasets $\mathcal{D}_1, \mathcal{D}_2,\mathcal{D}_3, \mathcal{D}_4$ by splitting the classes of the ImageNet-21K dataset~\cite{imagenet}. We create the splits based on the Wordnet~\cite{wordnet} hierarchy such that each $\mathcal{D}_i$ contains images from semantically similar classes. For each class, we hold out 25 images per class for evaluation.
The training data stream is created by sampling images at random from the four datasets $\{\mathcal{D}_{p_1}, \mathcal{D}_{p_2}, \mathcal{D}_{p_3}, \mathcal{D}_{p_4}\}$ where $[p_1, p_2, p_3, p_4]$ is a permutation of the sequence $[1,2,3,4]$. Images are sampled from the datasets sequentially such that images from $\mathcal{D}_{p_i}$ are seen only after most images of $\mathcal{D}_{p_{i-1}}$ are sampled (see Appendix for a detailed description of the sampling procedure), simulating a smooth change in semantic distribution. The goal is to learn a representation that can discriminate concepts from all datasets without overfitting or forgetting concepts seen earlier.

\subsection{Experiments with non-stationary distributions}

We train representations using conventional SimSiam, SimSiam with replay buffers (\cref{sec:buffer-ssl}) and SimSiam with minimum redundancy buffers (\cref{sec:minred}) on a single pass of this stream of data. During evaluation, we train a linear classifier on the learned representations to recognize all classes in the ImageNet-21k dataset, and measure the accuracy on the held-out set of each $\mathcal{D}_{p_i}$ separately. All results were averaged over 3 permutations of $p_1,\ldots, p_4$.

\cref{fig:continual-forgetting} plots the drop in classification accuracy on each dataset $D_{p_i}$ after the representation is trained on new data $D_{p_{i+1}}, D_{p_{i+2}},$ \etc, relative to the initial accuracy at the end of training on $D_{p_i}$. This serves as a measure of forgetting - a larger drop indicates that the representation is losing its ability to discriminate older classes. The results show that all methods suffer from forgetting. However, SimSiam with MinRed buffers displays less forgetting compared to conventional and buffered SimSiam. Intuitively, this can be attributed to the MinRed criteria that leads to retention of images from the older semantic distributions. \cref{fig:continual-accuracy} also shows the accuracy on all classes as training progresses. We observe that SimSiam with MinRed buffers consistently yields better generalization. 
In supplementary material, we also evaluated the learned representations on unseen classes, by testing only on future data streams $D_{p_{i+t}}$. Since MinRed buffers maintain training buffers with wider coverage of semantics, the learned representations were also shown to be more generalizable even to unseen concepts.

\section{Discussion and Future Work}
\label{sec:summary}

In this work, we exposed three challenges that require investigation to build robust deployable self-supervised learners. 
We improve the efficiency of Continuous SSL by leveraging replay buffers to revisit old samples. In future work, developing approaches for quickly rejecting samples by preemptively evaluating their value might yield improved data efficiency.
We also propose a novel minimum redundancy buffer to discard correlated samples allowing us to mimic the generation of IID training data, even in highly correlated settings. 
An alternative future direction could focus on learning representations that take advantage of the correlated nature of the data stream to learn from fine-grained discrepancies.

In data streams with non-stationary semantic distributions, we show that MinRed buffers alleviate the issue of catastrophic forgetting, as they are capable of maintaining unique samples from past distributions. 
However, we observed signs of saturating generalization as new concepts are introduced. Some possible reasons could be: 1) the cosine decay learning rate schedule and 2) the fixed capacity of our models that prohibits learning a large sequence of novel concepts. In preliminary experiments (see supplementary material), we saw that training with a constant learning rate (on 100M images from Flickr) does not lead to significant improvements in performance. We also observed that trivially expanding the architecture at regular intervals does not lead to noticeable improvements. However, we believe that further exploration in this direction is required to continually learning novel concepts in a self-supervised manner.

\section{Conclusion}
\label{sec:conclusion}

One of the grand goals of self-supervised learning is to build systems capable of continually learning from unlimited sources of unlabelled data. 
However, due to the need for benchmarking, existing SSL methods have primarily focused on curated datasets of limited size. Unfortunately, while the existing approaches work well in the dataset setup, we are still not close to deployable continual self-supervised methods. 
In this work, we advocate for a more realistic SSL setup that will facilitate deployment, while retaining the benefits of benchmarking. To this end, we identified three broad challenges of deployable SSL - training efficiency, correlated data, and lifelong learning, - and proposed potential solutions to address them. We believe however that further research is needed to develop deployable systems that deliver on the promise of self-supervised learning, and hope future efforts in SSL research focus on these challenges.

\bibliographystyle{splncs04}
\bibliography{refs}

\appendix
\section{Ethics and Negative Societal Impact}
\label{sec:ethics}

Research on self-supervised learning has been making progress towards building systems that can continually learn from data in our world without human supervision. In this work, we propose a problem setup that evaluates the challenges faced when such methods are truly deployed in-the-wild. As these systems start being deployed without supervision, there are numerous possibilities for biases to emerge based on the statistics of data consumed. These biases could potentially have a negative impact on our society. Therefore, it is important to exercise caution when deploying such systems and relying on them for downstream applications. Before deploying such systems it is also important to thoroughly study and implement approaches to mitigate such emergent biases. 

In our work, apart from working with existing datasets, we gather a collection of 100M images by downloading images from Flickr that have the Creative Commons license. While this license permits usage in our application, we do not plan to redistribute the images since they have not been thoroughly scanned for privacy concerns. The models trained in this work also have not been examined for potential societal biases or other spurious correlations that might have emerged from the data. While we plan to release the models trained here for research purposes, we would strongly advise against using them for any real-world applications.
\section{Implementation details}
\subsection{SimSiam}
The experiments conducted in this paper make extensive use of SimSiam~\cite{simsiam}, a contrastive learning algorithm for self-supervised visual representation learning. We closely follow the official SimSiam implementation available at \url{https://github.com/facebookresearch/simsiam}.

In all experiments, we used ResNet-18 with synchronized batch-norm as the backbone. All models were trained on 4 GPUs using stochastic gradient descent (SGD) with a batch size of $256$, learning rate of $0.05$ with a cosine decay, momentum of $0.9$, and weight decay of $0.0001$.
LIFO buffers are updated by always removing the oldest images. MinRed buffers are maintained by removing the most redundant images. Pseudo-code for the Buffered SSL with MinRed buffers is provided in \cref{alg:minred}.

\begin{algorithm}[t!]
\caption{Buffered SSL with MinRed buffer. PyTorch pseudo-code.}\label{alg:minred}
\begin{lstlisting}[language=Python]
def train(f, SimSiam, stream, num_updates):
  B = []  # Init empty buffer
  for ims in stream: # Load batch from stream
    Add2Buffer(B, ims)
    
    # Hyper-sampling: Update num_updates times
    for _ in range(num_updates):
      # Sample batch from buffer
      x = RandomSample(B)
      x1, x2 = aug(x), aug(x)
      z1, z2 = f(x1), f(x2)
        
      # Track features
      TrackRepresentations(B, x, (z1+z2)/2)
        
      # Compute loss and update models
      L = SimSiam(z1, z2)
      L.backward()        # Back-propagation
      update(f, SimSiam)  # SGD update
    
def Add2Buffer(B, ims):
  n_excess = len(B) + len(ims) - maxlen(B)
  if n_excess > 0:  # If full, remove n_excess.
    for _ in range(n_excess):
      # Pairwise dist
      d = pdist(B.feat, B.feat) 
      
      # Distance to nearest neig
      d_nneig = d.min(dim=1)
      
      # Remove sample with smallest d_nneig
      i_redundant = d_nneig.argmin(dim=0)
      B.remove(i_redundant)
  
  # Add new images to buffer
  for x in ims:
    B.add(x)
  
def TrackRepresentations(B, x, z, alpha=0.5):
  # EMA update
  B.feat[x] = alpha*B.feat[x] + (1 - alpha)*z
\end{lstlisting} 
\end{algorithm}

We follow the same evaluation protocol as in~\cite{simsiam,benchmarking}, and use linear probes on the learned features to recognize classes of three datasets.
The linear probes on ImageNet and iNaturalist were trained using the entirety of the datasets. On ImageNet, we trained the linear classifier using SGD+LARS~\cite{you2017large} for $10$ epochs with a batch size of $1024$, learning rate of $3.0$ with cosine decay, momentum of $0.9$, and no weight decay. On iNaturalist, the classifier was trained for $20$ epochs with learning rate of $12.0$.
Contrarily to ImageNet and iNaturalist, evaluations conducted on the full ImageNet dataset ($14M$ images) only used a subset of the data (using only 50 images per class for training and 25 for evaluation).
We trained the linear probe on this data using SGD+LARS for $30$ epochs and a learning rate of $3.0$.

\subsection{Sampling and Splits for Lifelong Learning}
In Section 6 of the main text, we construct a dataset with non-stationary semantic distributions to evaluate lifelong learning \textit{i.e.} learning without forgetting. Here we describe the process of construction of this dataset.

First, we performed a depth-first search (DFS) on the Wordnet\cite{wordnet} hierarchy. We split the sequence of DFS nodes (or classes) uniformly into four groups. Each such group contains classes that are close to each other in the Wordnet hierarchy and hence, semantically similar. In order to create the sequence of samples for lifelong learning, we could sequentially sample images from one split after the other. However, for future approaches, such hard boundaries in the sequence as we move from one split to the other could be easy to trivially identify and leverage to minimize forgetting. To make the setup more realistic, we create a smooth transition between one split to the other. The smooth transition is created by mixing the last 10\% of each split with the first 10\% of the next split. More concretely, we linearly decrease the likelihood of sampling from the first split and linearly increase the likelihood of sampling from the second split. 
\section{Additional results}

\begin{figure}[t!]
    \centering
    \includegraphics[width=\linewidth]{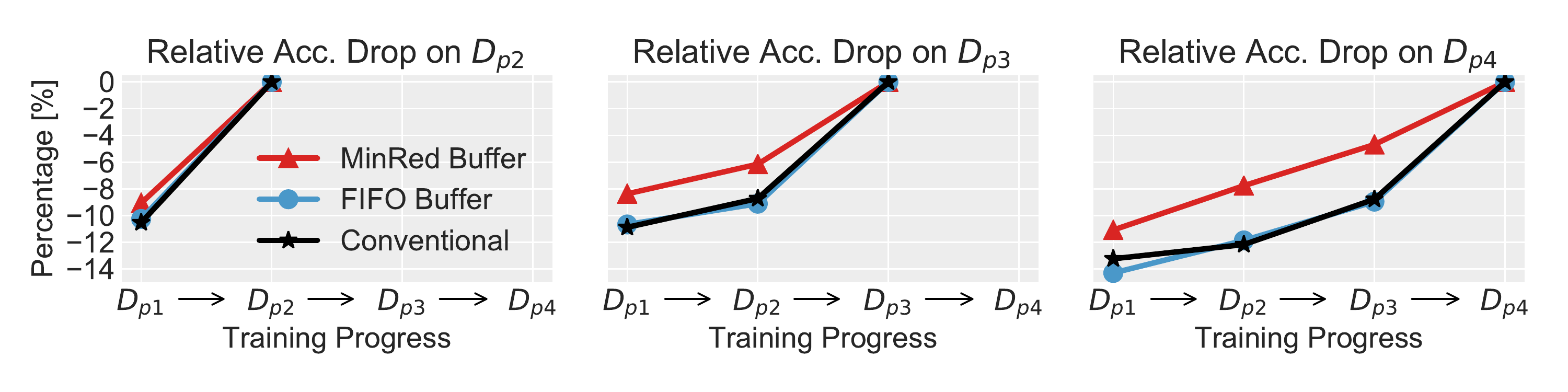}
    \caption{Open set generalization. While training on the data stream used for assessing continual learning, we also evaluated the models on future data partitions, which contain images from images from categories not yet seen during training. By training models with MinRed buffers, we can learn representations the can better generalize to unseen categories.}
    \label{fig:continual-openset}
\end{figure}

\subsection{Generalization towards unseen categories}
To assess the open set generalization ability of models trained with Minimum Redundancy (MinRed) buffers, we extended the continual learning experiment described in Section 6.2 and Figure 7 of the main paper, and further evaluate on future data partitions, \ie, data partitions containing categories yet unseen in the training sequence. The results are shown in Fig.~\ref{fig:continual-openset}. Training models with MinRed buffers also lead to better generalizable towards unseen categories. This is likely explained by the fact that MinRed buffers maintain higher semantic diversity in the training data, which encourages the model to learn more general representations, likely to generalize better to unseen categories.

\begin{figure}[t!]
    \centering
    \includegraphics[width=\linewidth]{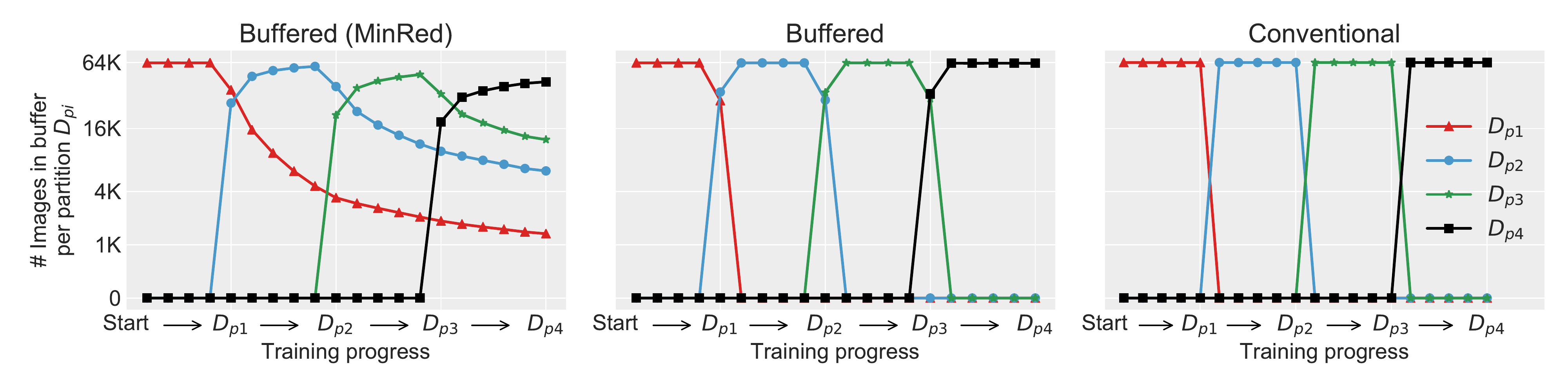}
    \caption{Contents of 64K buffers as the data distribution shifts over the course of training. For Conventional SSL (which has no buffer), we count images in a sequence of training batches totaling the same 64K images. Buffered SSL with a Minimum Redundancy (MinRed) buffer retains a significant number of images from previous data distributions. This is in contrast to Buffered SSL with LIFO buffers or Conventional SSL which have no ability to retain images for long periods of time.}
    \label{fig:continual-corr}
\end{figure}

\subsection{Buffer contents during lifelong learning}
To understand why MinRed buffers allow SimSiam to learn from non-stationary distributions with less forgetting (Section 6 of the paper), we analysed the contents of the buffer used to generate training samples. \cref{fig:continual-corr} shows the number of images in the buffer from each of the $D_{p1},\ldots,D_{p4}$ partitions, as training progresses from $D_{p1}$ to $D_{p4}$. As can be seen, only MinRed buffers are capable of retaining images from prior distributions. Since these images can then be sampled for training, MinRed buffers enable continual training with less forgetting.

\subsection{Learning rate schedules for continual learning}
\begin{figure}[t!]
    \centering
    \includegraphics[width=0.8\linewidth]{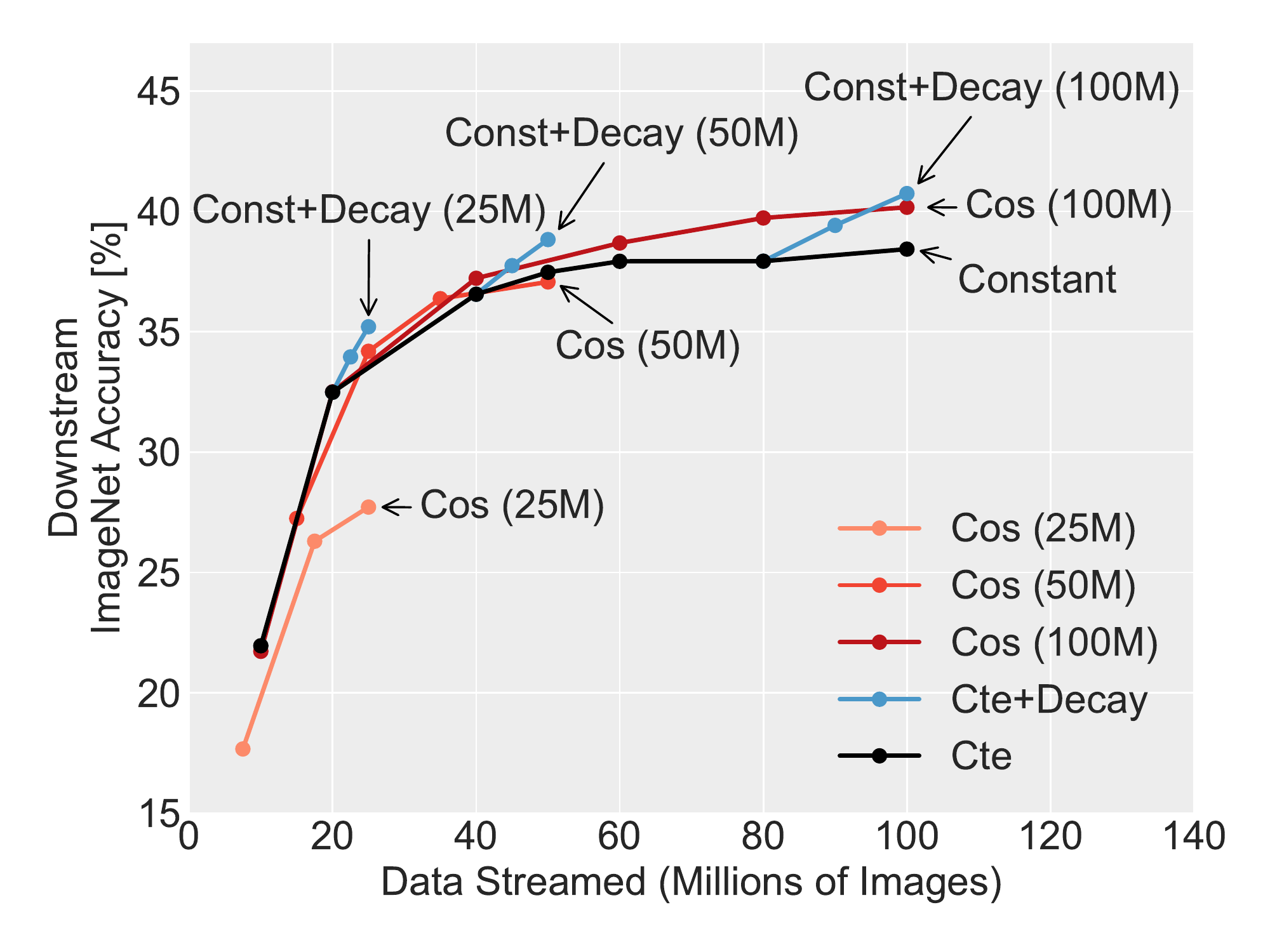}
    \caption{Downstream performance on ImageNet throughout self-supervised training with various learning rate schedules. ``Cos (xM)'' stands for cosine decay ending at iteration $x$/batch size. ``Const+Decay (xM)'' represents a learning rate schedule with a constant start (for about 80\% of the total training time), followed by a short cosine decay for the remainder of training.}
    \label{fig:lr_exp}
\end{figure}

The cosine learning rate schedule is not applicable to continuous SSL, as it requires a pre-determined end. We tested several learning rate schedules. Results are shown in \cref{fig:lr_exp}.
With a simple constant learning rate, models can still learn from a continuous (non-stationary) data stream, while still being able to achieve similar performances in the static case, when combined with a short learning rate decay before evaluation.

\end{document}